\def\year{2021}\relax
\documentclass[letterpaper]{article} % DO NOT CHANGE THIS
\usepackage{aaai21}  % DO NOT CHANGE THIS
\usepackage{times}  % DO NOT CHANGE THIS
\usepackage{helvet} % DO NOT CHANGE THIS
\usepackage{courier}  % DO NOT CHANGE THIS
\usepackage[hyphens]{url}  % DO NOT CHANGE THIS
\usepackage{graphicx} % DO NOT CHANGE THIS
\usepackage{natbib}  % DO NOT CHANGE THIS AND DO NOT ADD ANY OPTIONS TO IT
\usepackage{caption} % DO NOT CHANGE THIS AND DO NOT ADD ANY OPTIONS TO IT
\usepackage{booktabs}
\usepackage{multirow}
\title{Multi-Scale Spatial Temporal Graph Convolutional Network for Skeleton-Based Action Recognition}
\author{

    Zhan Chen,\textsuperscript{\rm 1}
	Sicheng Li,\textsuperscript{\rm 2}
	Bing Yang,\textsuperscript{\rm 1} 
	Qinghan Li,\textsuperscript{\rm 3}
	Hong Liu\textsuperscript{\rm 1}\thanks{Corresponding author}
\\
}
\affiliations{
    \textsuperscript{\rm 1}Key Laboratory of Machine Perception, Shenzhen Graduate School, Peking University, China\\
	\textsuperscript{\rm 2}Institute of Information and Communication Engineering, Zhejiang University, China\\
	\textsuperscript{\rm 3}Department of Electrical Engineering and Computer Science, Syracuse University, USA\\
	
	zhanchen\_cz@pku.edu.cn, jasonlisicheng@zju.edu.cn, bingyang@sz.pku.edu.cn, Qli160@syr.edu, hongliu@pku.edu.cn

}
\begin{document}

\maketitle

\begin{abstract}
	Graph convolutional networks have been widely used for skeleton-based action recognition due to their excellent modeling ability of non-Euclidean data.
	As the graph convolution is a local operation, it can only utilize the short-range joint dependencies and short-term trajectory but fails to directly model the distant joints relations and long-range temporal information that are vital to distinguishing various actions.
	To solve this problem, we present a multi-scale spatial graph convolution (MS-GC) module and a multi-scale temporal graph convolution (MT-GC) module to enrich the receptive field of the model in spatial and temporal dimensions.
	Concretely, the MS-GC and MT-GC modules decompose the corresponding local graph convolution into a set of sub-graph convolution, forming a hierarchical residual architecture. 
	Without introducing additional parameters, the features will be processed with a series of sub-graph convolutions, and each node could complete multiple spatial and temporal aggregations with its neighborhoods.
	The final equivalent receptive field is accordingly enlarged, which is capable of capturing both short- and long-range dependencies in spatial and temporal domains.
	By coupling these two modules as a basic block, we further propose a multi-scale spatial temporal graph convolutional network (MST-GCN), which stacks multiple blocks to learn effective motion representations for action recognition.
	The proposed MST-GCN achieves remarkable performance on three challenging benchmark datasets, NTU RGB+D, NTU-120 RGB+D and Kinetics-Skeleton, for skeleton-based action recognition.
\end{abstract}

\section{Introduction}
Human action recognition has received a lot of attention in various applications, such as video surveillance, human-machine interaction, virtual reality, video analysis and so on  \citep{poppe2010survey,weinland2011survey,aggarwal2011human,sudha2017approaches}.
The current studies on human action recognition are mainly divided into RGB video based and skeleton based.
Compared with RGB video based methods which require high-cost computing resources to process pixel-level information in RGB image or temporal optical flow \citep{tran2015learning,simonyan2014two,zhang2016real,li2020tea},
skeleton based methods are more computationally efficient as skeleton data represents a human structure with 2D or 3D coordinates of a few dozen joints. 
Moreover, skeleton data conveys relatively high-level information for human actions, which exhibits stronger robustness to appearance variation and environmental noises, $e.g.$, background clutter, illumination changes \citep{wang2018rgb,vemulapalli2014human,liu2017enhanced,tu2018skeleton}.
Therefore, human action recognition with skeleton data has been attracting increasing attention.

\begin{figure}[t]
	\centering
	\includegraphics[width=0.95\columnwidth]{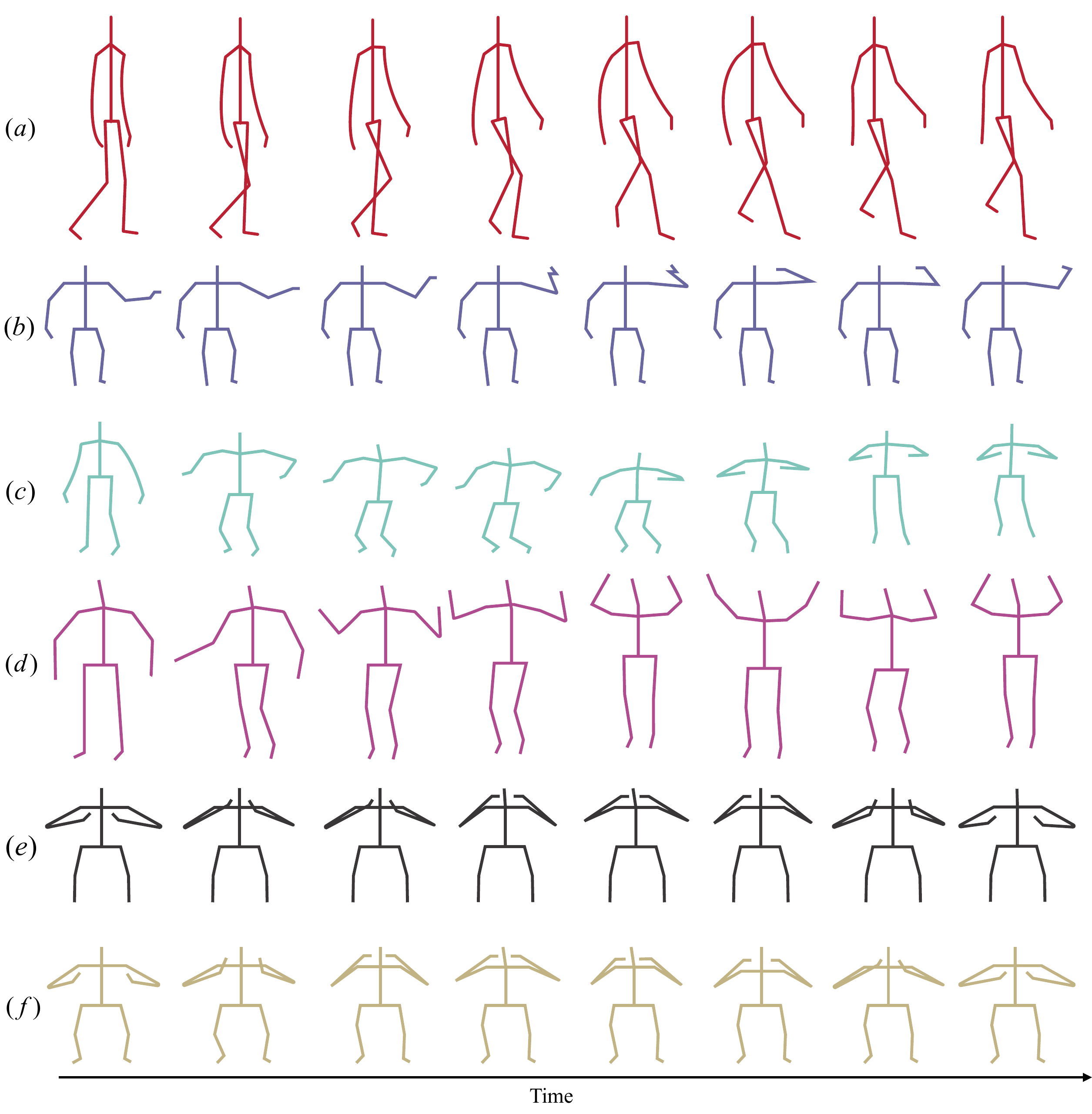} % Reduce the figure size so that it is slightly narrower than the column. Don't use precise values for figure width.This setup will avoid overfull boxes.
	\caption{Some samples from NTU RGB+D dataset. (a) ``walking'' (b) ``hand waving'' (c) ``jumping up'' (d) ``cheering up'' (e) ``wearing on glasses'' (f) ``taking off glasses''}.
	\label{motivation}
\end{figure}

To extract discriminative spatial and temporal features for skeleton-based action recognition, great efforts have been made to learn patterns embedded in the spatial configuration of the joints and their temporal dynamics.
Conventional deep-learning-based methods learn human action representations with convolutional neural networks (CNNs) and recurrent neural networks (RNNs).
These methods manually construct the skeleton sequence as a kind of grid-shape structure such as a pseudo-image \citep{ke2017new,soo2017interpretable,li2017skeleton} or a sequence of the coordinate vectors \citep{du2015hierarchical,zhang2017view,liu2016spatio}.
As the skeleton is naturally structured as a graph in non-Euclidean geometric space, these methods fail to fully explore the inherent relationships between human joints. 
Recently, Graph convolutional network (GCN) -based method \citep{yan2018spatial} addressed these drawbacks and constructed a spatial temporal skeleton graph with the joints as graph nodes and natural connectivities in both human body structures and time as graph edges, therefore, the spatial temporal relations between human joints are well embedded in the adjacency matrix of the skeleton graph.

The follow up works \citep{shi2019skeleton2, shi2019skeleton, si2019attention} have proposed many effective variants of GCN, they mostly consider the short-range connections.
However, long-range dependencies are also important for action recognition.
For instance, as shown in Fig. \ref{motivation}, (1) ``Walking'' requires coordination of the whole body to maintain balance, while ``hand waving'' can be accomplished just by hand.
Since different actions require coordination of different body parts to complete, it is of critical importance to design a multi-scale feature extractor to capture the dependencies between different range joints.
(2) ``Jumping up'' and ``cheering up'' could be rather ambiguous when only considering the local movements of the leg parts.
Therefore, capturing the essential contextual information of an action which usually requires a large receptive field could help us to judge whether the jump is just a ``jumping up'' or a jump for ``cheering up''.
(3) ``Wearing on glasses'' and ``taking off glasses'' are very ambiguous in a short period, which requires the algorithm to capture long-range temporal information and reduce ambiguity.
In summary, capturing both short-range joints dependencies and distant joints relations in the spatial domain and considering both short-term trajectory and long-range temporal information are vital to skeleton-based action recognition.
There are many GCN-based works considering one or both of those aspects \citep{li2019actional,shi2019two,zhang2020context,peng2020learning,liu2020disentangling,li2019symbiotic}.
Nevertheless, the problem is far from being solved.

To capture both short-range joints dependencies and distant joints relations, %existing methods achieve this by higher-order polynomials of the skeleton adjacency \citep{li2019actional} leading to increased parameters and complexity.
we proposed a simple yet efficient multi-scale spatial graph convolution (MS-GC) module which is capable of extracting multi-scale spatial features.
The MS-GC replaces the spatial graph convolution operation with a group of sub-graph convolutions which formulate a hierarchical structure with residual connections between adjacent subsets.
When the features go through the module, they realize multiple information exchanges with neighboring nodes, and the equivalent spatial receptive field is thus increased to capture the relations between distant joints.
For multi-scale temporal modeling, %the parallel local temporal convolution kernel is usually used to enrich the temporal receptive field of the model \citep{liu2020disentangling}.
we naturally extend the MS-GC to the temporal domain and propose a multi-scale temporal graph convolution (MT-GC) module. 
With a similar architecture as the MS-GC module, the MT-GC module is capable of modeling the long-range temporal relationship over distant frames.
These two modules are complementary and cooperate in endowing the network with both multi-scale spatial and multi-scale temporal modeling abilities.
We devise two approaches to combine these two modules as a basic block and propose a multi-scale spatial temporal graph convolution network (MST-GCN) by stacking multiple blocks.
Experiments on three large-scale datasets for skeleton-based action recognition verify the effectiveness of the obtained model: when the parameters of MST-GCN are less than one third of ST-GCN, our model perform much better than ST-GCN, and when the model has parameters equivalent to ST-GCN, it achieves remarkable results compared to the state-of-the-art methods\footnote{\url{https://github.com/czhaneva/MST-GCN}}.
Our contributions can be summarized as follows:
\begin{quote} 
	\begin{itemize}
		\item A multi-scale spatial graph convolution module to capture both local joints connectives and non-local joints relations for spatial modeling.
		\item A multi-scale temporal graph convolution module to efficiently enlarge the temporal receptive field for long-range temporal dynamics.
		\item Integrating the MS-GC and MT-GC modules, we propose a multi-scale spatial temporal graph convolutional network. 
		On three datasets for skeleton-based action recognition, our method achieves remarkable results with the same parameters as ST-GCN.
	\end{itemize}
\end{quote}

\section{Related Works}

\subsubsection{Neural Networks with Graph.}
Graph neural network has attracted a lot of attention due to the effective representation for graph structure data such as social networks and the human skeleton.
Existing GCN-models are mainly divided into two flows: The spectral-based method and the spatial-based method.
The spectral-based GCNs \citep{defferrard2016convolutional,kipf2016semi} transform graph signals to graph spectral domains based on eigen-decomposition, and perform convolution in the graph spectral domain, which is limited in terms of computational efficiency due to the heavy computation cost brought by the eigen-decomposition.
In contrast to the spectral-based GCN, the spatial-based GCN \citep{simonyan2014two,niepert2016learning} is a generalization of the convolutional network from Euclidean space to non-Euclidean space, which directly implements convolution operation on the graph nodes and their neighbors in the spatial domain. 
This work follows the spatial-based method.

\subsubsection{Skeleton-Based Action Recognition.}
As an alternative data source for RGB video, skeleton data has been widely used for action recognition due to the robustness against variations of backgrounds.
The methods for skeleton-based human action recognition are mainly divided into two flows: the hand-crafted-based and the deep-learning-based.
The hand-crafted-based methods usually focus on designing handcrafted features based on the physical intuitions such as joints angles \citep{ofli2014sequence}, distances \citep{xia2012view} and kinematic features \citep{zanfir2013moving}.
However, manually designing features requires a lot of energy, and it is difficult to consider all factors related to the action. 
With the development of deep learning, data-driven approaches which automatically learn the action patterns have attracted much attention.
Some RNN-based models model the skeleton data as time sequences of coordinate vectors and capture temporal dynamics between consecutive frames, such as bi-RNNs \citep{du2015hierarchical}, deep LSTMs \citep{liu2016spatio,shahroudy2016ntu}, and attention-based model \citep{song2016end}.
CNN-based models which try to embed the motion patterns into a skeleton image also achieve remarkable results \citep{ke2017new,soo2017interpretable}.

Recently, the graph-based methods draw much attention due to their superiority to model the relations between body joints.
\citeauthor{yan2018spatial} proposed a spatial temporal graph convolutional network (ST-GCN) to directly model the skeleton data as graph structure \citep{yan2018spatial}.
To capture the relations between distant joints, some data-dependent methods were proposed.
\citeauthor{shi2019two, li2019actional, zhang2020context} introduced additional mechanism to adaptively learn the relation between different joints \citep{shi2019two, li2019actional, zhang2020context}.
On the other hand, some approaches extracted multi-scale structural features via higher-order polynomials of the skeleton adjacency matrix.
\citeauthor{li2019actional, peng2020learning, liu2020disentangling} introduced multiple-hop modules to break the limitation of representational capacity caused by one-order approximation \citep{li2019actional, peng2020learning, liu2020disentangling}.
Different from their methods, we utilize a series of sub-graph convolutions cascaded by residual connections to capture both short-range joints dependencies and distant joints relations.
For multi-scale temporal modeling,
\citeauthor{liu2020disentangling} enhanced the regular temporal modeling with multi-scale learning and deployed parallel $3 \times 1$ kernel sizes with different dilation rates to enlarge the temporal receptive fields \citep{liu2020disentangling},
while our model enriches the temporal receptive field within a single block and aggregates both short- and long-range temporal information via a hierarchical architecture.

\section{Methodology}
In this section, we first briefly review ST-GCN \citep{yan2018spatial} to make the paper self-contained.
Then, we detail our MST-GCN for skeleton-based action recognition.
\subsection{Basic Graph Convolution}
A human skeleton graph is defined as $\mathcal{G} = \{\mathcal{V},\mathcal{E}\}$,
where $\mathcal{V} = \{v_{ti}|t=1, \cdots,T, i=1,\cdots,V\}$ includes all the joints in a skeleton sequence and $\mathcal{E} = \{\mathcal{E}_{\mathcal{S}}, \mathcal{E}_{\mathcal{T}}\}$ is the set of intra-skeleton and inter-frames edges. $T$ denotes the number of frames and $V$ denotes the number of nodes in a human skeleton.

In the spatial domain, a sampling function is defined as $\mathcal{B}_{\mathcal{S}}(v_{ti}) = \{v_{tj}|d(v_{tj}, v_{ti}) \leq D\}$ to determine the region range of the spatial graph convolution.
Here $d(v_{tj}, v_{ti})$ denotes the shortest path from $v_{tj}$ to $v_{ti}$, the parameter $D$ controls the spatial range to be included in the neighbor graph. 
The direct natural connections in human skeleton can be represented by the spatial adjacency matrix $A_{\mathcal{S}} \in R^{V \times V}$, where $A_{\mathcal{S}_{ij}} = 1$ if the joint $i$ and $j$ are directly connected and 0 otherwise.
To capture refined location information, a spatial configuration labeling function is proposed in ST-GCN \citep{yan2018spatial}.
It divides the neighbor set into three subsets: (1) the root node itself; (2) the centripetal subset, which contains the neighboring nodes that are closer to the center of gravity than the root node; (3) the centrifugal subset, which contains the neighboring nodes that are farther from the center of gravity than the root node.
Thus $A_{\mathcal{S}}$ is accordingly parted to be $A_{\mathcal{S}}^{root}$, $A_{\mathcal{S}}^{centripetal}$ and $A_{\mathcal{S}}^{centrifugal}$.
We have 3 different subsets $\mathcal{P} = \{root, centripetal, centrifugal\}$.

For the temporal dimension,  the sampling function is defined as $\mathcal{B}_{\mathcal{T}}(v_{ti}) = \{v_{qi}| \left| q-t \right| \leq \lfloor \Gamma/2 \rfloor \}$, where parameter $\Gamma$ controls the temporal scales to aggregate.
The temporal adjacency matrix is defined as $A_{\mathcal{T}} \in R^{T \times T}$ to represent the trajectories of the joints between consecutive frames.
With a simple labeling function based on the temporal sequence, the temporal adjacency matrix can be partitioned into $\Gamma$ parts: $A_{\mathcal{T}}^{- \lfloor \Gamma/2 \rfloor}, \cdots, A_{\mathcal{T}}^{0}, \cdots, A_{\mathcal{T}}^{ \lfloor \Gamma/2 \rfloor}$.

With the defined graph, refined sampling and labeling function, the graph convolution operations on node $v_{ti}$ are formulated as:
\begin{equation}\label{SGC}
Y_{\mathcal{S}}(v_{ti}) = \sum_{v_{tj} \in \mathcal{B}_{\mathcal{S}}(v_{ti})}\frac{1}{Z_{ti}(v_{tj})}X(v_{tj})W_{\mathcal{S}}(\mathcal{L}_{\mathcal{S}}(v_{tj})),
\end{equation}
\begin{equation}\label{TGC}
Y_{\mathcal{T}}(v_{ti}) = \sum_{v_{tj} \in \mathcal{B}_{\mathcal{T}}(v_{ti})}\frac{1}{Z_{ti}(v_{qi})}X(v_{qi})W_{\mathcal{T}}(\mathcal{L}_{\mathcal{T}}(v_{qi})),
\end{equation}
where, $X(\cdot)$ is the feature of a node. $W_\mathcal{S}(\cdot)$ and $W_\mathcal{T}(\cdot)$ are weight functions that allocate a weight indexed by label $\mathcal{L}_{\mathcal{S}}(v_{tj})$ and $\mathcal{L}_{\mathcal{T}}(v_{qi})$ from a set of weights.
$Z_{ti}(\cdot)$ is the number of the corresponding subset, which normalizes feature representations.
$Y_{\mathcal{S}}(v_{ti})$ and $Y_{\mathcal{T}}(v_{ti})$ denote the outputs of spatial graph convolution and temporal graph convolution at node $v_{ti}$, respectively.

\subsection{Implementation}
The implementation of the spatial graph convolution is not straightforward.
Specifically, we denote the feature map of the network is $X \in R^{C \times T \times V}$, where $C$ denotes the number of channels.
To implement the ST-GCN \citep{yan2018spatial}, Eq. \ref{SGC} is transformed into 
\begin{equation}\label{RE-SGC}
Y_{\mathcal{S}} = \sum_{p \in \mathcal{P}} W_{\mathcal{S}}^{(p)}(X\hat{A}_{\mathcal{S}}^{(p)}) \odot M^{(p)},
\end{equation}
where, $\hat{A}_{\mathcal{S}}^{(p)} = D_{\mathcal{S}}^{(p)^{-\frac{1}{2}}}A_{\mathcal{S}}^{(p)}D_{\mathcal{S}}^{(p)^{\frac{1}{2}}}$  is a normalized adjacency matrix in spatial configuration of the $p$ subset, 
and $D_{\mathcal{S}}^{(p)}=\sum_{j}A_{\mathcal{S}_{ij}}^{(p)} + \alpha$ is a degree matrix. 
$\alpha$ is set to 0.001 to avoid empty rows.
$W_{\mathcal{S}}^{(p)} \in R^{C_{out} \times C_{in} \times 1 \times 1}$ is a trainable weight vector of the $1 \times 1$ convolution operation.
$M^{(p)} \in R^{N \times N}$ is a mask map that indicates the importance of each joint. 
$\odot$ denotes the dot product.
As mentioned in \citep{shi2019two,shi2019skeleton2}, the dot multiplication operation can change only the importance of existing edges without adding new edges.
Therefore, we replace the dot product operation with an additional operation, the Eq. \ref{RE-SGC} is transformed into
\begin{equation}\label{RE-ASGC}
Y_{\mathcal{S}} = \sum_{p \in \mathcal{P}} W_{\mathcal{S}}^{(p)}X(\hat{A}_{\mathcal{S}}^{(p)} + M^{(p)}),
\end{equation}

Considering the consistency of the spatial temporal graph and the video in temporal dimension, we implement temporal graph convolution with classical convolution operation, and Eq. \ref{TGC} can be represented as:
\begin{equation}
Y_{\mathcal{T}} = Conv2D[K_t \times 1](X),
\end{equation}
where $Conv2D[K_{t} \times 1]$ is the classic 2D convolutional operation with the kernel size $K_{t} \times 1$.

In ST-GCN, $D$ is set to 1 and $K_{t}$ is set to 9, which results in the limited spatial temporal receptive field of the model and lack of flexibility.
To solve these problems, we propose MS-GC and MT-GC modules.

\subsection{Multi-Scale Spatial Graph Convolution Module}
\begin{figure}[t]
	\centering
	\includegraphics[height=0.6\textwidth]{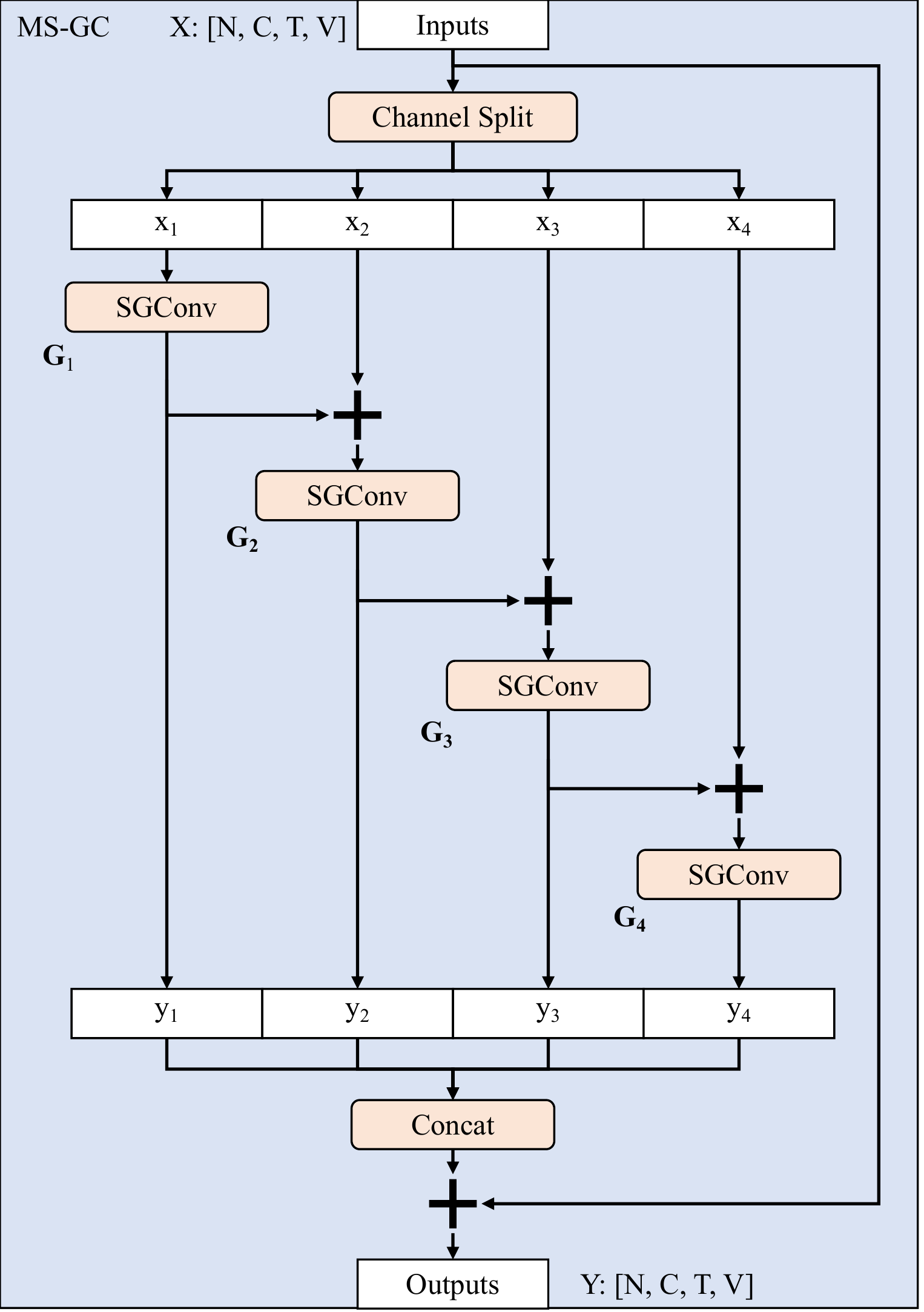} % Reduce the figure size so that it is slightly narrower than the column. Don't use precise values for figure width.This setup will avoid overfull boxes.
	\caption{Illustration of multi-scale spatial graph convolution (MS-GC) module. SGConv denotes the spatial graph convolution, $N$ is the batch size.} 
	\label{MS-GC}
\end{figure}

%Many human actions require coordination between parts that are far apart, such as action ``wearing a shoe'' requires the foot and hand to move collaboratively.
%
Recent studies show that capturing dependencies between both short-range nodes and long-range nodes is essential for skeleton based human action recognition \citep{li2019actional,shi2019two,zhang2020context,liu2020disentangling}.
However, these methods either introduce additional modules to adaptively learn the relationship between different nodes or use higher-order polynomials of the skeleton adjacency matrix.
They fail to well control the balance between performance and parameters.
In this paper, we propose the MS-GC module, in which the spatial features and corresponding local graph convolution are split into a group of subsets.
The main idea is inspired by Res2Net \citep{gao2019res2net} which has a great impact on many fields of image processing.
We transfer it to GCN-based action recognition with skeleton data.
The architecture of the proposed MS-GC module is shown in Fig. \ref{MS-GC}.
The subsets are formulated as a hierarchical residual architecture, so features can be hierarchically processed.
The equivalent receptive field of the spatial dimension is enlarged and the model is capable of capturing the relations between those distant joints.

Specifically, given an input feature $X$ whose shape is $[C, T, V]$, we split the feature into $s$ fragments along the channel dimension, denoted by $\mathbf{x}_{i}$, where $i \in \{1,2,\cdots,s\}$.
The shape of each fragment thus becomes $[C/s, T, V]$.
Each fragment $\mathbf{x}_{i}$ has a corresponding spatial graph convolution which is implemented using Eq. \ref{RE-ASGC}, denoted by $\mathbf{G}_{i}$.
Each sub-spatial graph convolution has $1/s$ number of channels compared with the original ones and accordingly has only $1/s^{2}$ parameters as the original ones.
Moreover, the residual connection is placed between two adjacent fragments, which enrich the diversity of receptive fields to capture the dependencies between both local and non-local joints.
Formally,
\begin{equation}
\mathbf{y}_{i}=\left\{\begin{array}{ll}
			\mathbf{G}_{i}(\mathbf{x}_{i}) & i=1 \\
			\mathbf{G}_{i}(\mathbf{x}_{i} + \mathbf{y}_{i-1}) & i > 1 
			 \end{array}
			\right.
\end{equation}
where $\mathbf{y}_{i} \in \mathbf{R}^{C/s \times T \times V}$ is the output of $i_{th}$ sub-spatial graph convolution.

In this module, these $s$ fragments have different receptive fields.
For instance, with $D$ set to 1, $\mathbf{G}_{1}$ can aggregate information from 1-hop neighbors,
while $\mathbf{G}_{2}$ could potentially receive feature information from 2-hop neighbors by aggregating the information from $\mathbf{y}_{1}$.
Thus the equivalent receptive field of the last fragment $\mathbf{y}_{s}$ has been enlarged several times.
Finally, all the fragments are concatenated and an additional residual connection for the entire module is adopted to help the model converge.
The outputs of MS-GC module can be computed as:
\begin{equation}
Y = \sigma([\mathbf{y}_{1};\cdots;\mathbf{y}_{s}] + X)
\end{equation}
where $\sigma$ is the activation function. 
The obtained output feature $Y$ involves capturing spatial feature representations between nodes at different distances which is superior to the local spatial representations obtained by using a single local graph convolution in typical approaches.

The MS-GC module is efficient: it can control the balance between complexity and the multi-scale representation ability of the model by adjusting $s$.
Without introducing additional parameters and time-consuming operations, the module is able to capture the dependencies between both short- and long-range joints.

\subsection{Multi-Scale Temporal Graph Convolution Module}
Temporal modeling is essential to action recognition as well, while long-range temporal modeling is largely ignored in previous works.
Long-range temporal dependence can not only reduce the ambiguity between actions through temporal information, but also provide global information to help the model better learn spatial temporal features.
Many existing works perform temporal modeling using temporal convolutions with a fixed kernel size throughout the architecture \citep{yan2018spatial,li2019actional,shi2019two,peng2020learning}.
The long-range temporal relationship is indirectly modeled by repeated stacking local temporal graph convolutions in deep networks.
However, after a large number of local convolution operations, the useful features from distant frames have already been weakened and cannot be captured well.

To address this problem described above, we naturally extend the MS-GC module to the temporal domain.
The proposed MT-GC module has a similar structure to MS-GC, but does not introduce additional residual connections.
It replaces the local temporal graph convolution with a set of sub-temporal graph convolutions which are constructed as hierarchical residual-like connections.
The temporal graph convolution in each subset is the same, but has different inputs.
Similar to the MS-GC module, when the spatial temporal features go through the MT-GC module, a series of cascaded temporal graph convolution operations are applied to corresponding fragments to enlarge the temporal receptive field.
Finally, a simple concatenation strategy is adopted to fuse the fragments.
Therefore, the final output has the multi-scale temporal representation while both the short-range and long-range temporal relationships are well captured.

\subsection{Multi-Scale Spatial Temporal Graph Convolution}
\begin{figure}[t]
	\centering
	\includegraphics[height=0.9\columnwidth]{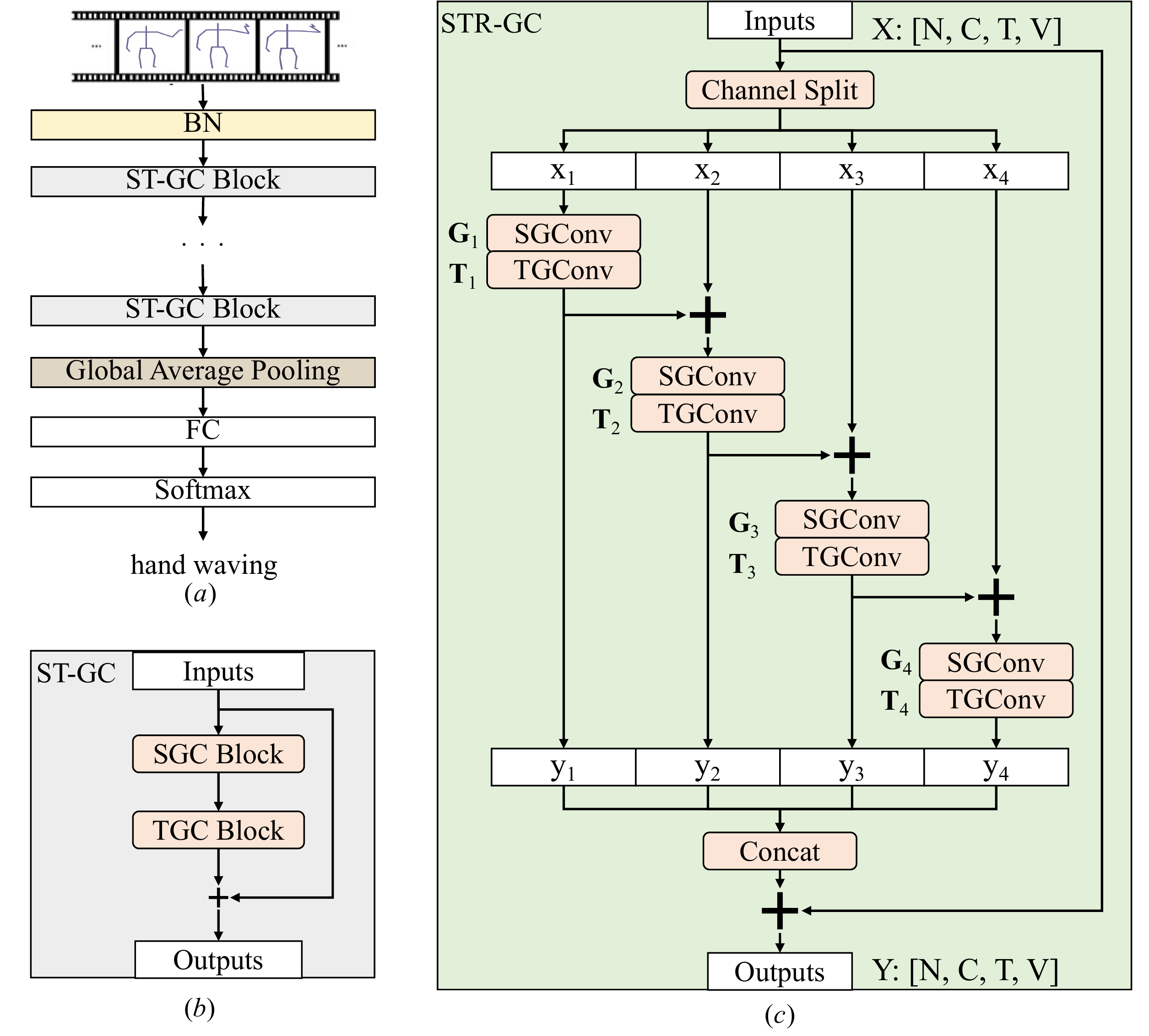} % Reduce the figure size so that it is slightly narrower than the column. Don't use precise values for figure width.This setup will avoid overfull boxes.
	\caption{Architecture overview. (a) The full architecture of the ST-GCN backbone, BN is a batch normalization layer, FC is a full connected layer. (b) Illustration of ST-GC block which is used to build the whole network. (c) Illustration of proposed spatial temporal residual graph convolution (STR-GC) module, TGConv is the temporal graph convolution operation.} 
	\label{overview}
\end{figure}

To have a fair comparison with the state-of-the-art methods, we follow the same method as ST-GCN \citep{yan2018spatial} to construct our multi-scale spatial temporal graph convolution network.
The ST-GCN backbone is shown as Fig. \ref{overview}(a), which consists of 1 batch normalization layer to normalize the data and 10 ST-GC blocks to extract spatial and temporal features.
Each ST-GC block, illustrated as Fig. \ref{overview}(b), contains a spatial graph convolution and a followed temporal graph convolution to extract spatial and temporal features alternately.
A global average pooling operation is adopted to aggregate the spatial temporal information and the final prediction results are output through the full connected layer together with a softmax layer. 

Our work focuses on extracting discriminative spatial and temporal features, and there are two modes of combining the MS-GC module with the MT-GC module.
In the first method, we replace the spatial graph convolution unit with our MS-GC module,
and the temporal graph convolution is replaced with our MT-GC module as well.
In the second method, we construct a spatial temporal residual graph convolution (STR-GC) module by concatenating the sub- spatial and temporal graph convolution within a single block, which is illustrated as Fig. \ref{overview}(c).
For convenience, we denote the sub-temporal graph convolution as $\mathbf{T}_{i}$.
In this module, the spatial and temporal features will be updated alternately in each subset, and the corresponding spatial and temporal receptive field are enlarged due to the combined explosion effect.
Formally,
\begin{equation}
\mathbf{y}_{i}=\left\{\begin{array}{ll}
\mathbf{T}_{i}(\mathbf{G}_{i}(\mathbf{x}_{i})) & i=1 \\
\mathbf{T}_{i}(\mathbf{G}_{i}(\mathbf{x}_{i} + \mathbf{y}_{i-1})) & i > 1 
\end{array}
\right.
\end{equation}

The first module can maintain the continuity of the model in extracting multi-scale temporal feature and spatial feature,
while the latter is lighter than the first one, and it is more convenient to expand the spatial temporal joint learning that we leave to future work.
The comparison of these two modules will be involved in the ablation study.

\section{Experiments}
\subsection{Datasets}

\subsubsection{NTU RGB+D:}NTU RGB+D \citep{shahroudy2016ntu} dataset contains 60 different human action classes and it is the most widely used dataset for evaluating skeleton-based action recognition models.
It consists of 56,880 action samples in total which are performed by 40 distinct subjects.
The 3D skeleton data is collected by Microsoft Kinect v2 from three cameras simultaneously with different horizontal angles: -45$^{\circ}$, 0$^{\circ}$, 45$^{\circ}$.
The human pose in each frame is represented by 25 joints.
There are two evaluation protocols for this dataset: Cross-Subject (X-sub) and Cross-View (X-view).
Under the former protocols, half of the 40 subjects consists of the training set and the other for testing.
For the latter, samples captured by camera 2 and 3 are used for training and the rest are for testing.

\subsubsection{NTU-120 RGB+D:}NTU-120 RGB+D \citep{Liu_2019_NTURGBD120} dataset is the currently largest 3D skeleton-based action recognition dataset captured under various environmental conditions.
This dataset is an expansion of NTU RGB+D dataset in the number of performer and action categories.
It contains 114,480 action samples from 120 action classes.
Samples are captured by 106 subjects with three camera views.
There are 32 setups that denote a specific location and background.
Following the recommendation of the author, two benchmarks are used to evaluate the performance of models.
Cross-Subject (X-sub) split the 106 subjects into training and testing group. Each group contains 53 subjects.
Cross-setup (X-setup) divides samples with even setup IDs for training and odds setup IDs for testing.

\subsubsection{Kinetics-Skeleton:}Kinetics dataset \citep{kay2017kinetics} contains about 300,000 video clips in 400 classes collected from the Internet. 
The skeleton information is not provided by the original dataset but estimated by publicly available OpenPose \citep{cao2017realtime} toolbox.
The captured skeleton information contains 18 body joints, along with their 2D coordinates and confidence score.
There are 240,436 samples for training and 19794 samples for testing.
Following the conventional evaluation method, Top-1 and Top-5 accuracies are reported.

\subsubsection{Training details}
The model is built by stacking 10 building blocks, the first four blocks of the baseline model have 64 channels for output. 
In the 5$_{th}$ and 8$_{th}$ blocks, the number of channels is doubled while the temporal length is downsampled by a factor 2 according to the previous blocks.
In the comparison experiment, the number of channels is adjusted to ensure a fair comparison under the same level of parameter complexity.
We use SGD with Nesterov momentum (0.9), batch size 24, an initial learning rate 0.1 (can linearly scale up with batch size \citep{goyal2017accurate}) to train the model for 110 epochs.
The learning rate is decayed by 10 at 50$_{th}$, 70$_{th}$ and 90$_{th}$ epoch.
For the frame which contains more than two persons, only the top-2 persons are selected based on the average energy.
All skeleton sequences are padded to $T=$300 frames by replaying the actions, but a sliding window of size 150 is used to randomly choose the  crop  input of sequence for Kinetics-Skeleton dataset. 
The same data pre-processing as in \citep{shi2019two} is adopted for NTU RGB+D and NTU-120 RGB+D datasets.
All experiments in the ablation study used the above setting, including both our proposed method and the baseline model.

\subsection{Ablation Study}
In this subsection, we examine the effectiveness of the proposed MS-GC module, MT-GC module and their combination.
The number of channels for each block will be adjusted to maintain the consistency of the parameters. 
The baseline model is the adaptive ST-GCN which is discussed in Sec. 3.2 implementation. 
\subsubsection{MS-GC module} To verify the effectiveness of the proposed MS-GC module, we compare the MS-GCN with adaptive ST-GCN and perform the comparison experiments on NTU RGB+D with the benchmark of cross-subject.
To build our MS-GCN, we replace the SGC block in 9 residual ST-GC blocks with our MS-GC module.
The only difference between our MS-GC module and the SGC block is the set of $s$. 
When $s$ set to 1, the MS-GC module is actually the SGC block, and when $s > 1$, the module gains the ability of multi-scale spatial representation.
Here we first compare the number of the split subsets $s$ with different settings.
As shown in Table \ref{table:MS-GC}, with the increase of $s$, the MS-GCN achieves better performance with the similar parameters.
Specifically, when $s$ set to 4, our MS-GC module can improve the baseline at 1.0\%, indicating that the importance of capturing the dependencies between distant joints and that our proposed MS-GC module can effectively enlarge the receptive field to capture such dependencies.
To further explore the effectiveness of the MS-GC module, we also designed a lightweight MS-GCN to compare with other models.
Each block of the lightweight MS-GCN has the same number of channels as ST-GCN, but the entire network has less parameters.
Comparing the lightweight MS-GCN with the baseline model, our model achieves better performance.
The improved performance demonstrates that our proposed MS-GC module is equipped with the ability to capture the relations between the local and non-local joints for better spatial representations.

\begin{table}[h]
	\centering
	\begin{tabular}{l|cc|c}
		\toprule
			  &Setting	       &Params	 &Top 1 \\
		\midrule
		ST-GCN&64$c$ $\times$ 1$s$ &3.1M     &87.2\% \\
		\midrule
		\multirow{4}*{MS-GCN}&34$c$ $\times$ 2$s$ &3.1M	 &87.8\% \\
						 	 &23$c$ $\times$ 3$s$ &3.0M	 &88.1\% \\
							 &17$c$ $\times$ 4$s$ &3.0M	 &\textbf{88.2\%} \\
							 &16$c$ $\times$ 4$s$ &\textbf{2.7M}	 &88.0\% \\
		\bottomrule
	\end{tabular}
	\caption
	{
		Comparisons between regular SGC and MS-GCN modules with difference of $s$ on NTU RGB+D with X-sub benchmark. $c$ denotes the output channels of the sub-SGCs in the second block. $s$ denotes the number of subsets we divide the spatial features into.
	}
	\label{table:MS-GC}
\end{table}

\subsubsection{MT-GC module}
To validate the efficacy of MT-GC module, we also compare our MT-GCN with the adaptive ST-GCN on NTU RGB+D cross-subject benchmark.
We build our MT-GCN by replacing the TGC of ST-GCN with MT-GC module, and the only difference between our MT-GCN and the baseline is inserting our MT-GC module.
Here we first evaluate the effectiveness and efficiency of different $s$.
As shown in Table \ref{table:MT-GC}, when the number of split subsets is increased, we observe consistently better results with similar parameters, verifying MT-GC module is capable of capturing long-range temporal dynamics that are rarely considered in current work.
Especially, when we split the inputs into 4 subsets along with the channel dimensions, we improve the accuracy by 1.2\%.
We also designed a lightweight MT-GCN which has the same channels setting for each block as the ST-GCN.
Compared with the baseline model, the lightweight MT-GCN outperforms the ST-GCN with an absolute improvement of 0.7\%, which demonstrates that our MT-GCN is efficient.
All the above comparison experiments have demonstrated the effectiveness of our proposed module.
It has the ability to extract both short- and long-range temporal dependencies.

\begin{table}[t]
	\centering
	\begin{tabular}{l|cc|c}
		\toprule
		&Setting	       &Params	 &Top 1 \\
		\midrule
		ST-GCN&64$c$ $\times$ 1$s$ &3.1M     &87.2\% \\
		\midrule
		\multirow{4}*{MT-GCN}&40$c$ $\times$ 2$s$ &3.1M	 &87.7\% \\
							 &30$c$ $\times$ 3$s$ &3.1M	 &88.1\% \\
							 &24$c$ $\times$ 4$s$ &3.1M	 &\textbf{88.4\%} \\
							 &16$c$ $\times$ 4$s$ &\textbf{1.4M}	 &87.9\% \\
		\bottomrule
	\end{tabular}
	\caption
	{
	Comparisons between regular TGC and MT-GCN modules with difference of $s$ on NTU RGB+D with X-sub benchmark. $c$ is the output channels of the sub-TGCs in the second block. $s$ is the number of subsets we divide the temporal features into.
	}
	\label{table:MT-GC}
\end{table}

\subsubsection{MST-GC module} 
We conduct multi-scale spatial and temporal graph convolution and explore method to effectively combine the two proposed modules.
According to previous ablation experiments, $s$ is set to 4 for both MS-GC and MT-GC modules to achieve the best performance.
The MST-GCN replaces both SGC and TGC in the ST-GC block with the MS-GC and MT-GC modules, respectively.
While the STR-GCN replaces ST-GC blocks with the STR-GC module.
As shown in Table \ref{table:MST-GC}, the MS-GC and MT-GC modules are complementary and their combination can promote each other to achieve better performance.
STR-GCN is a model with potential, it achieves good performance with fewer parameters under the same settings.
MST-GCN improves the adaptive ST-GCN at 1.8\% with the similar parameters and 0.9\% with only one third parameters, which validates the complementarity of these two module and also demonstrates that our MST-GCN can capture multi-scale spatial and temporal dependencies, leading to better motion patterns learning.

\begin{table}[h]
	\centering
	\small
	\tabcolsep1.55mm
	\begin{tabular}{c|c|cc|c}
		\toprule
		Spatial model& Temporal model &Setting &Params	 &Top 1 \\
		\midrule
		Regular SGC& Regular TGC&64$c$ $\times$ 1$s$ &3.1M     &87.2\% \\
		MS-GC& Regular TGC&17$c$ $\times$ 4$s$ &3.1M	 &88.2\% \\
		Regular SGC&MT-GC&24$c$ $\times$ 4$s$ &3.1M	 &88.4\% \\
		MS-GC&MT-GC&16$c$ $\times$ 4$s$ &\textbf{0.9M}	 &88.1\% \\
		MS-GC&MT-GC&30$c$ $\times$ 4$s$ &3.0M	 &\textbf{89.0\%} \\
		\multicolumn{2}{c|}{STR-GC}&30$c$ $\times$ 4$s$ &2.8M	 &88.5\% \\
		\bottomrule
	\end{tabular}
	\caption
	{
		The effectiveness and efficiency of MST-GCN. The accuracy is on NTU RGB+D cross-subject task. $s$ is the number of subsets we divide both the spatial temporal features into. 
	}
	\label{table:MST-GC}
\end{table}

\subsubsection{Feature visualization.} 
\begin{figure}[t]
	\centering
	\includegraphics[height=0.6\columnwidth]{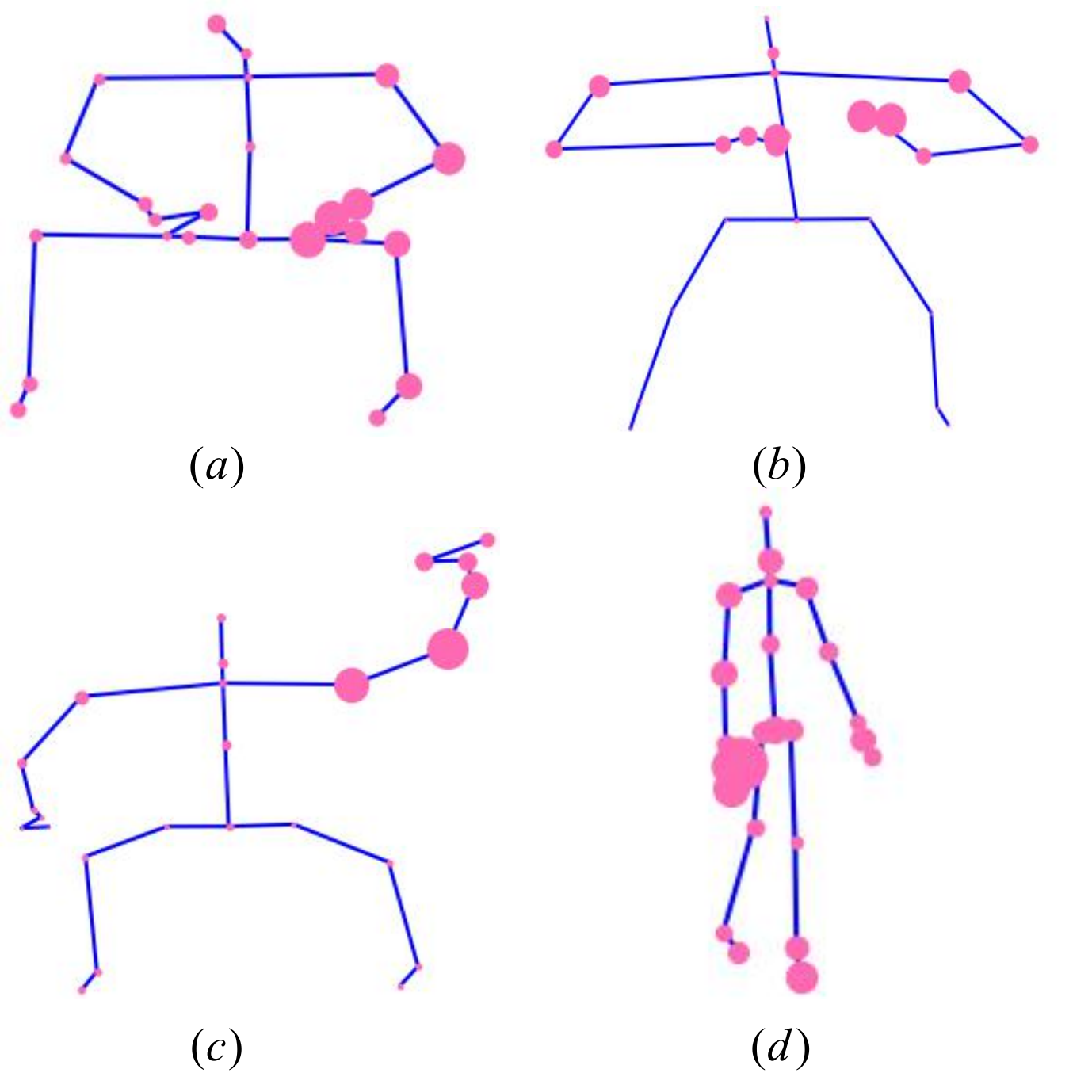} % Reduce the figure size so that it is slightly narrower than the column. Don't use precise values for figure width.This setup will avoid overfull boxes.
	\caption{Feature responses in the last layer of MST-GCN backbone. The areas of circles indicate the response magnitudes. (a) ``wearing a shoe'' (b) ``clapping'' (c) ``hand waving'' (d) ``walking''} 
	\label{feature}
\end{figure}

To validate how the features of each joint affect the final performance, we visualize the output feature maps of the last MST-GC block in Fig. \ref{feature},
where the circle around each joint indicates magnitude of feature responses of this joint.
As shown in Fig. \ref{feature}, our model is capable of focusing on the parts that are most relevant to the action.
Specifically, both the left arm and left foot among action ``wearing a shoe'' is well focused.
Among action ``clapping'' and ``hand waving'', large responses are distributed on the arm parts, and there is almost no response from other parts, which could improve the robustness of the model and reduce the interference of noise nodes.
The model could also capture the coordination of the whole body when walking.
On the other hand, our MST-GCN could capture useful long-range dependencies to recognize the actions.

\subsection{Comparison with State of the Arts}
Considering the complementary information between different data, many state-of-the-art methods utilize multi-stream fusion strategies. 
To conduct a fair comparison, we adopt the same multi-stream fusion strategy as \citet{shi2019skeleton2,shi2019skeleton,cheng2020skeleton}, which utilizes the joint, bone, joint motion and bone motion streams.
The ``joint stream'' uses the original skeleton coordinates as input, and the ``bone stream'' uses the the differential of spatial coordinates (the second order information of skeleton joints) as input.
The ``joint motion stream'' and ``bone motion stream'' uses the temporal differential between adjacent frames of corresponding data as input. 
A simple score-level fusion strategy is adopted to obtain the fused score.

\begin{table}[t]
	\centering
	\small
	\tabcolsep1.70mm
	\begin{tabular}{lcc}
		\toprule
		Methods&								X-view (\%)&		X-sub (\%)	\\
		\midrule
		HBRNN \citep{du2015hierarchical}&			   64.0&		59.1	\\
		P-LSTM \citep{shahroudy2016ntu}&		       67.3&		60.7	\\
		TCN \citep{soo2017interpretable}&		       83.1&		74.3	\\
		VA-LSTM \citep{zhang2017view}&		           87.7&		79.2	\\
		%	3scale ResNet 152 \citep{li2017skeleton}&		84.6&		90.9	\\
		\midrule
		ST-GCN \citep{yan2018spatial}&			       88.3&		81.5	\\
		AS-GCN \citep{li2019actional}&			       94.2&		86.8	\\
		%	PB-GCN \citep{thakkar2018part}&			87.5&		93.2	\\
		2s AGC-LSTM \citep{si2019attention}&		   95.0&		89.2	\\
		2s AGCN \citep{shi2019two}&		               95.1&		88.5	\\
		2s NAS-GCN \citep{peng2020learning}&		   95.7&		89.4	\\
		4s DGNN \citep{shi2019skeleton2}&			   96.1&		89.9	\\
		4s MS-AAGCN \citep{shi2019skeleton}&		   96.2&		90.0	\\
		2s MS-G3D \citep{liu2020disentangling}&		   96.2&	\textbf{91.5}\\
		4s Shift-GCN \citep{cheng2020skeleton}&	       96.5&		90.7	\\
		\midrule
		Js MST-GCN (ours)&							   95.1&	    89.0	 \\	
		Bs MST-GCN (ours)&							   95.2&	    89.5	 \\	
		2s MST-GCN (ours)&							   96.4&	    91.1	 \\	
		4s MST-GCN (ours)&						\textbf{96.6}&	\textbf{91.5}	 \\		
		\bottomrule
	\end{tabular}
	\caption
	{
		Comparisons of the Top-1 accuracy with the state-of-the-art methods on the NTU RGB+D dataset.
	}
	\label{table:NTU60}
\end{table}

\begin{table}[!h]
	\centering
	\small
	\tabcolsep0.75mm
	\begin{tabular}{lcc}
		\toprule
		Methods&		X-sub (\%)&		X-setup (\%)	\\
		\midrule
		ST-LSTM \citep{liu2016spatio}&	55.7&		57.9	\\
		MTCNN + RotClips \citep{ke2018learning}&	62.2&		61.8	\\
		SkeMotion \citep{liu2017global}&	67.7&		66.9	\\
		TSRJI \citep{caetano2019skeleton}&	67.9&		62.8	\\
		\midrule
		4s Shift-GCN \citep{cheng2020skeleton}&			85.9&		87.6	\\
		2s MS-G3D \citep{liu2020disentangling}&	86.9&		88.4	\\
		\midrule
		Js MST-GCN (ours)&		82.8&	84.5	 \\	
		Bs MST-GCN (ours)&		84.8&	86.3	 \\	
		2s MST-GCN (ours)&		87.0&	88.3	 \\	
		4s MST-GCN (ours)&		\textbf{87.5}&	\textbf{88.8}	 \\		
		\bottomrule
	\end{tabular}
	\caption
	{
		Comparisons of the Top-1 accuracy with the state-of-the-art methods on the NTU-120 RGB+D dataset.
	}
	\label{table:NTU120}
\end{table}

\begin{table}[!h]
	\centering
	\small
	\tabcolsep1.70mm
	\begin{tabular}{lcc}
		\toprule
		Methods&		Top-1 (\%)&		Top-5 (\%)	\\
		\midrule
		ST-GCN \citep{yan2018spatial}&		31.6&		53.7	\\
		AS-GCN \citep{li2019actional}&		34.8&		56.3	\\
		2s AGCN \citep{shi2019two}&	36.1&		58.7	\\
		4s DGNN \citep{shi2019skeleton2}&		36.9&		56.9	\\
		2s NAS-GCN \citep{peng2020learning}&	37.1&		60.1	\\
		2s MS-G3D \citep{liu2020disentangling}&		38.0&		\textbf{60.9}	\\
		\midrule
		Js MST-GCN (ours)&		36.0&	58.5	 \\	
		Bs MST-GCN (ours)&		32.3&	58.2	 \\	
		2s MST-GCN (ours)&		37.8&	60.3	 \\	
		4s MST-GCN (ours)&		\textbf{38.1}&	60.8	 \\		
		\bottomrule
	\end{tabular}
	\caption
	{
		Comparisons of the Top-1 and Top-5 accuracy with the state-of-the-art methods on the Kinetics dataset.
	}
	\label{table:Kinetics}
\end{table}

The results of 1-stream which only uses the joint or bone stream and 2-stream which utilizes the both joint and bone stream and 4-stream which uses all 4 streams are reported in our experiments.
To verify the generality of our approach, we compare the MST-GCN with state-of-the-art methods on three different datasets: NTU RGB+D dataset, NTU-120 RGB+D dataset and Kinetics-Skeleton dataset,
the corresponding results are shown in Table \ref{table:NTU60}, Table \ref{table:NTU120}, Table \ref{table:Kinetics}.
On NTU RGB+D dataset, MST-GCN achieves competitive performance on cross-view and cross-subject benchmarks.
For NTU-120 RGB+D dataset, MST-GCN achieves the-state-of-the-art results on both cross-subject and cross-setup benchmarks.
For Kinetics-skeleton, MST-GCN achieves the-state-of-the-art result on Top-1 and competitive result on Top-5.
The competitive and the state-of-the-art results verify the superiority of our model.

\subsection{Discussion}
As discussed in ST-GCN \cite{yan2018spatial}, the skeletons can provide complementary information to RGB-based and optical flow-based models.
We also explore using MST-GCN to capture motion information in two-stream style action recognition. 
The RGB model is a standard temporal segment networks (TSN) \cite{wang2016temporal} with pretrained weights\footnote{https://github.com/open-mmlab/mmaction2}.
For fair comparison, both ST-GCN and MST-GCN only utilize the joint stream.
As shown in Table \ref{table: Ensemble}, there is a big gap between the RGB-based models and the skeleton-based models, but the skeleton-based models can provide useful motion information to boost the performance of the RGB-based models.
Adding MST-GCN to the RGB-based model leads to 1.7\% increase, much better than ST-GCN (0.8\%).
These results clearly show that the improvement obtained in skeleton-based models could further improve the performance of the ensemble model. 

\begin{table}[t]
	\centering
	\small
	\tabcolsep1.1mm
	\begin{tabular}{c|ccc|c}
		\toprule
				    & RGB TSN & ST-GCN  & MST-GCN	 &Top 1 (\%) \\
		\midrule
		\multirow{3}*{Single Model}     & $\surd$ &         &            &70.4       \\
		       &         & $\surd$ &     	     &31.6       \\
	                &         &         & $\surd$    &36.0       \\
	    \midrule
		\multirow{2}*{Ensemble Model}    & $\surd$ & $\surd$ &            &71.2       \\
		       & $\surd$ &         & $\surd$    &72.1       \\
		\bottomrule
	\end{tabular}
	\caption
	{
		Combining RGB model with different skeleton models. The accuracy is on Kinetics validation set.
	}
	\label{table: Ensemble}
\end{table}

\section{Conclusion}
In this work, we propose a MS-GC module and a MT-GC module to capture both short- and long-range dependencies in spatial and temporal domains, respectively.
The MS-GC and MT-GC modules enhance the multi-scale representation ability of the model by decomposing a local convolution into several sub-convolutions and combining them through residual connections.
Both of them are simple but efficient, they are also complementary to each other.
By coupling these methods, we propose a multi-scale spatial and temporal graph convolution network (MST-GCN), which is a powerful network to extract multi-scale spatial and temporal features.
On three large-scale challenging datasets, the proposed MST-GCN achieves competitive or the-state-of-the-art results. 
Considering the relation between the spatial and temporal, jointly extracting spatial-temporal features is very helpful to explore the motion pattern in the spatial-temporal domain.
In the future, we will try the spatial and temporal joint learning to promote the performance of human action recognition.
We will further explore the promotion of the skeleton-based methods to RGB-based methods.

\section{Ackonwledgements}
This work was supported by National Natural Science Foundation of China (U1613209), Science and Technology Plan Projects of Shenzhen (JCYJ20190808182209321, JCYJ20200109140410340).

\bibliographystyle{aaai}
\bibliography{mybib}

\end{document}